\documentclass[conference]{IEEEtran}
\usepackage{hyperref}
\IEEEoverridecommandlockouts

\usepackage{cite}
\usepackage{amsmath,amssymb,amsfonts}
\usepackage{algorithmic}
\usepackage{graphicx}
\usepackage{textcomp}
\usepackage{url}
\usepackage{xcolor}
\def\BibTeX{{\rm B\kern-.05em{\sc i\kern-.025em b}\kern-.08em
    T\kern-.1667em\lower.7ex\hbox{E}\kern-.125emX}}
\begin{document}

\title{	PolyBERT: Fine-Tuned Poly Encoder BERT-Based Model for Word Sense Disambiguation\\
}

\author{\IEEEauthorblockN{Linhan Xia}
\IEEEauthorblockA{\textit{ICNLab, Shenzhen Graduate School, } \\
\textit{ Peking University,}\\
 Shenzhen, P.R.China \\
linhanxia@aliyun.com}
\and
\IEEEauthorblockN{Mingzhan Yang}
\IEEEauthorblockA{\textit{ICNLab, Shenzhen Graduate School,} \\
\textit{Peking University, }\\
Shenzhen, P.R.China \\
my47@illinois.edu}
\and
\IEEEauthorblockN{Guohui Yuan}
\IEEEauthorblockA{\textit{ICNLab, Shenzhen Graduate School, } \\
\textit{ Peking University,}\\
 Shenzhen, P.R.China \\
yuangh@pkusz.edu.cn}
\and
\IEEEauthorblockN{Shengnan Tao}
\IEEEauthorblockA{\textit{ICNLab, Shenzhen Graduate School, } \\
\textit{ Peking University,}\\
 Shenzhen, P.R.China \\
1021203565@qq.com}
\and
\IEEEauthorblockN{Yujing Qiu}
\IEEEauthorblockA{\textit{ICNLab, Shenzhen Graduate School, } \\
\textit{ Peking University,}\\
 Shenzhen, P.R.China \\
qyj62442@126.com}
\and
\IEEEauthorblockN{Guo Yu}
\IEEEauthorblockA{\textit{ CEC GienTech Technology Co.,Ltd. } \\
 Shenzhen, P.R.China \\
yu.guo18@gientech.com}
\and
\IEEEauthorblockN{Kai Lei*}
\IEEEauthorblockA{\textit{ICNLab, Shenzhen Graduate School, } \\
\textit{ Peking University,}\\
 Shenzhen, P.R.China \\
leik@pkusz.edu.cn, \\corresponding author}
}
\maketitle

\begin{abstract}
Mainstream Word Sense Disambiguation (WSD) approaches have employed BERT to extract semantics from both context and definitions of senses to determine the most suitable sense of a target word, achieving notable performance. However, there are two limitations in these approaches. First, previous studies failed to balance the representation of token-level (local) and sequence-level (global) semantics during feature extraction, leading to insufficient semantic representation and a performance bottleneck. Second, these approaches incorporated all possible senses of each target word during the training phase, leading to unnecessary computational costs. To overcome these limitations, this paper introduces a poly-encoder BERT-based model with batch contrastive learning for WSD, named PolyBERT. Compared with previous WSD methods, PolyBERT has two improvements: Firstly, (1) a poly-encoder with a multi-head attention mechanism is employed to integrate both token-level (local) and sequence-level (global) semantics, rather than focusing solely on one aspect. This approach enhances semantic representation by effectively balancing local and global semantics. Secondly, (2) to avoid redundant training inputs, Batch Contrastive Learning (BCL) is introduced. BCL utilizes the correct senses of other target words in the same batch as negative samples for the current target word, which reduces training inputs and computational cost. The experimental results demonstrate that PolyBERT outperforms baseline WSD methods such as Huang's GlossBERT and Blevins's BEM by 2\% in F1-score. In addition, PolyBERT with BCL reduces GPU hours by 37.6\% compared with PolyBERT without BCL.

\end{abstract}


\section{Introduction}
Word Sense Disambiguation (WSD) refers to determining the most suitable sense of target ambiguous word according to a specific context \cite{Navigli2009}. As a fundamental task in Natural Language Processing (NLP), WSD play a crucial role in various applications such as machine translation \cite{parameswarappa2011kannada}, information retrieval \cite{Zhong2012} and named entity recognition \cite{moro2014entity}.

A major limitation of mainstream WSD works is insufficient semantic representation due to imbalanced extraction of token-level (local) and sequence-level (global) semantics. Mainstream works are categorized into two types: sequence-based and token-based \cite{bevilacqua2021recent}. Sequence-based works neglect local semantics (of each token), while token-based works overlook global semantics (of the entire sequence).

Another limitation of mainstream WSD works is unnecessary computational cost caused by redundant training data inputs. WSD models learn feature differences between the correct sense and incorrect sense during training phase. For a specific target word, incorrect senses can be derived from the correct senses of other target words in the same batch. However, mainstream works input all candidate senses of each target word during the training phase, leading to significant redundancy.

To address the above-mentioned limitations, this paper proposes a poly-encoder BERT-based WSD model with batch contrastive learning (BCL), named PolyBERT. The contributions of this paper are following:
\begin{enumerate}
    \item To address the issue of imbalanced representation between token-level (local) and sequence-level (global) semantics, this paper introduces a poly-encoder-based feature extraction approach. This approach employs a multi-head attention mechanism to fuse extracted token-level (local) and sequence-level (global) semantics, integrating both token-level and sequence-level semantics into the representation, rather than focusing on just one.
    \item Prior works sampled all possible senses of each target word to generate positive and negative sample pairs, but this overlooked the potential to use positive samples from other target words in the same batch as negative samples. This oversight led to data redundancy and reduced computational efficiency. To address this, this paper adopts Batch Contrastive Learning, which leverages positive samples from other target words as negative samples, thereby avoiding the inefficiency of comprehensive negative sampling in prior works.
    \item This paper constructs a WSD model (PolyBERT) and numerous experiments are conducted to evaluate the performance of PolyBERT. The source code and trained models of PolyBERT is available at \href{https://github.com/Xia12121/PolyBERT\_for\_WSD}{here}.
\end{enumerate}

To evaluate the effectiveness of PolyBERT, experiments are conducted on public English-all-word datasets. According to the experiment results, compared to the best mainstream works (BEM \cite{blevins2020moving}), PolyBERT achieves a 2\% higher F1-score. In addition, strategy of contrastive learning saves 37.6\% GPU hours compared with learning strategy of mainstream works. 

In section \ref{section:Related Works}, this paper introduces related researches. In section \ref{section:Methodology}, this paper introduces the details of PolyBERT. In section \ref{section:Experiment}, this paper reports the experimental setup and results of the experiments. In section \ref{sec:conclusion} this paper summarizes our work.

\section{Related Works}\label{section:Related Works}
Pre-trained Transformer-based Language Models (PLMs) has been proven significant in improving performance of WSD. Vial et al. \cite{vial2019sense} initially employed BERT model to generate embedding, and used this embedding to disambiguate word sense. Since then, the PLMs-based WSD methods has become the mainstream. PLMs-based works can be divided into two categories: sequence-based approaches and token-based approaches \cite{bevilacqua2021recent}.

Sequence-based approaches disambiguate word sense based on semantics of whole sequence (global semantics). One of the most representative works is GlossBERT proposed by Huang et al. \cite{huang2019glossbert}. GlossBERT combined context and gloss as a sequence, and model utilize the sequence to determine whether gloss matches the sense of target word. Based on GlossBERT, Yap et al. \cite{yap2020adapting} enhance representation by introducing example of each sense from WordNet. Differ from GlossBERT, ESC proposed by Barba \cite{barba2021esc} combined context and all candidate glosses as a sequence. Based on the sequence model is trained to predict which gloss is the correct definition of correct sense. However, sequence-based approaches disambiguate word sense without extract semantic from each token especially token of target word (local semantics), which limit the capability of semantic representation.

Differ from sequence-based approaches, Token-based approaches employ semantics of each token (local semantics) to disambiguate word sense. Similar to PolyBERT, BEM proposed by Blevins et al. \cite{blevins2020moving} consists of two independent encoder to embed context and gloss of sense respectively. BEM extract the token of target word from context's embedding and token [CLS] from gloss's embedding as representations, which are used to determine the correct sense by dot product. Similar to BEM, Zhang et al. \cite{zhang2024quantum} employed two encoders. Moreover, Zhang et al. employed quantum interference to enhance representation of target word and glosses. However, token-based approaches only extract semantics from each token, which limit the capability of representing global semantics.

Existing works are not able to represent local and global semantics in a balance way, which cause insufficient representation of semantics and limitation of performance.

\section{Methodology}\label{section:Methodology}
\begin{figure*}[!t]
\centering
\includegraphics[width=\textwidth]{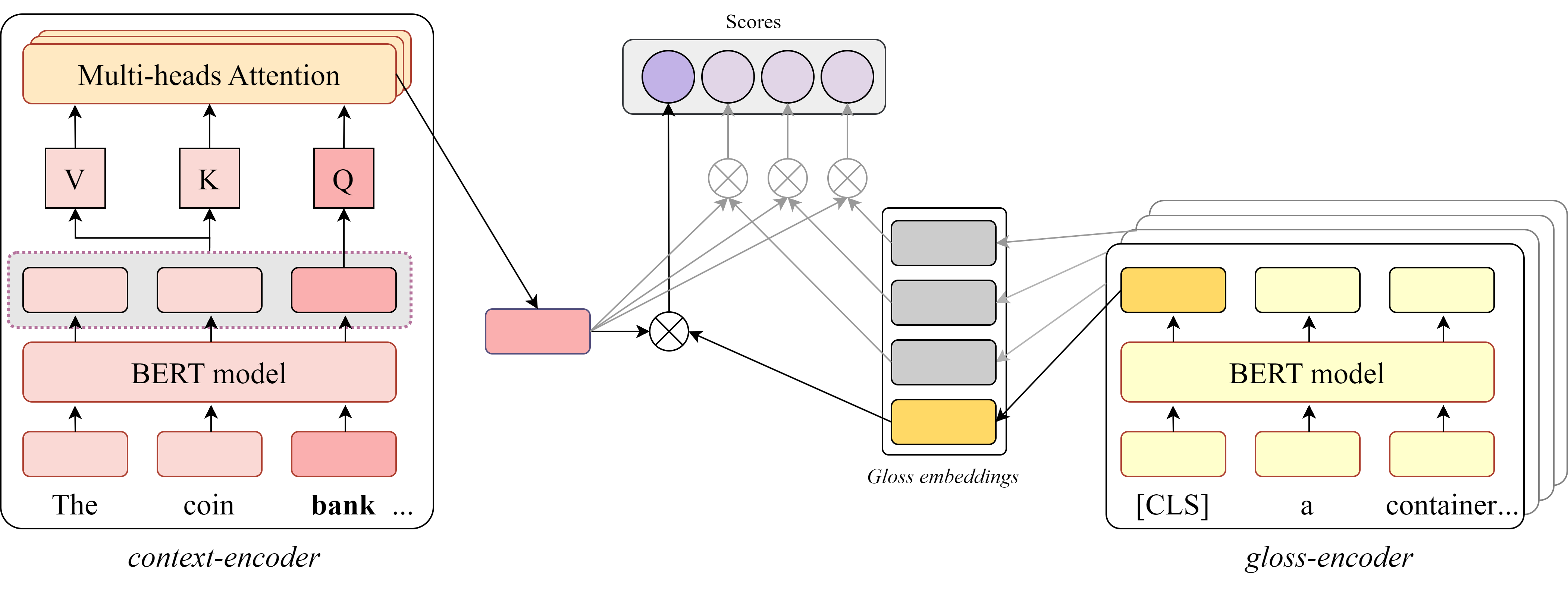}  
\caption{Pipeline of PolyBERT. The \textit{context-encoder} embeds target word ant its context, the \textit{gloss-encoder} embeds gloss of sense. We extract \(t^{th}\) token from output of context embedding, where \(w^t\) it the target word. \textit{Context-encoder} employs multi-head attention mechanism to fuse local and global semantics to generate representation of target word.  \textit{Gloss-encoder} takes Token [CLS] as representation of gloss. PolyBERT employs dot product to calculate the semantic similarities (scores) of each sense.}
\label{fig:pipeline}
\end{figure*}

This section illustrate the details of PolyBERT. PolyBERT consists of two independent encoders: (1) \textit{context-encoder}: aims to embed target word with its surrounding context and \textit{gloss-encoder}: aims to embed definition (gloss) of sense. PolyBERT fuses the embeddings from \textit{context-encoder} and \textit{gloss-encoder} to evaluate which sense is the most suitable. The pipeline of PolyBERT is shown in Fig. \ref{fig:pipeline}.

PolyBERT for WSD is divided into two phases: \textbf{batch contrastive pre-learning} (training phase) and \textbf{sense prediction}. During the batch contrastive pre-learning phase, PolyBERT only processes the gloss of the target word’s correct sense, which differs from prior works. During the sense prediction phase, PolyBERT intake target word and its all candidate definitions.

Following the introduction of the PolyBERT pipeline, this section outlines the notation employed in the model's description. Formally, the context \(c\) is composed of multiple words \(w\), with the \(t^{th}\) word designated as the target word \(w^t\). The WSD system is formulated as a function \(f(w^t,c)=s\), where \(s\in S_{w^t}\), the set of all possible senses of \(w^t\). Each sense \(s\) is associated with a gloss \(g\), where \(g\in G_{w^t}\). Within \(S_{w^t}\) and \(G_{w^t}\). \(s^c\) and its corresponding gloss \(g^c\) denote the correct sense and gloss of \(w^t\), as derived from context \(c\).

\subsection{Context-encoder and Gloss-encoder}\label{sec:3.1}
Encoders of PolyBERT aim to generate embeddings of textual sequences and extract representations from the embeddings. Each encoder of PolyBERT is initialized with BERT model (denoted as BERT). The BERT-based encoders embed sequences with BERT-specific start and end token: token [CLS] and token [SEP]. Therefore, the embedding can be read as \(E=T^{[CLS]},T^1,T^2,...,T^n,T^{[SEP]}\), where \(n\) is the length of input textual sequence and each token \(T\) represents a word \(w\) except \(T^{[CLS]}\) and \(T^{[SEP]}\).

\textit{Context-encoder} denoted as \(B_C\) is employed to embed target word with its context.  The input sequence of \(B_C\) can be written as \(c=w^1,...,w^t,...,w^n\), where the \(t^{th}\) word of \(c\) is the target word \(w^t\). The BERT model will embed the textual sequence into embedding sequence \(E_C\), and the representation \(r\) of target word read as: \begin{equation}
    r_{w^t}={E_C}[t]=B_C(c)[t],
    \label{eq:representation_tw}
\end{equation}where \(r_{w^t}\) is local semantics, and \(E_C\) is global semantics. To balance representation of local and global semantics, context-encoder adopt poly-encoder and multi-heads attention mechanism to fuse them. Firstly, context-encoder replicate \(r_{w^t}\) \(poly_m\) times to form query matrix \(Q\)
\begin{equation}
    Q = \mathbf{1}_{poly_m} \otimes r_{w^t}
    \label{eq:poly}
\end{equation}\(poly_m\) is a hyperparameter in the poly-encoder, representing the number of tokens that are generated after the extraction process. Secondly, context-encoder take the embedding sequence \(E_C\) as key \(K\) and value \(V\) matrix and fuse \(Q,K,V\) through multi-head attention mechanism. The calculation of each head read as

\begin{equation}
    head_i=\text{softmax}\left( \frac{(Q\cdot W_i^Q)(K\cdot W_i^K)}{\sqrt{d_k}} \right)\cdot V \cdot W_i^V,
    \label{eq:head}
\end{equation}where \(W_i^Q \in \mathbb{R}^{d_{\text{model}} \times d_k}\), \(W_i^K \in \mathbb{R}^{d_{\text{model}} \times d_k}\) and \(W_i^V \in \mathbb{R}^{d_{\text{model}} \times d_v}\) are linear projection matrices, \(d_k\) is dimension of \(K\) and \(d_v\) is dimension of \(V\). Then we concat each head to generate fused representation of local and global semantic information. The concat process read as
\begin{equation}
    r_{w^t}^F = Concat\left(head_1,\dots ,head_h \right)\cdot W^O,
    \label{eq:fuse}
\end{equation}where, \(h\) is the number of heads and \(W^O\) is concatenated linear projection matrix.

\textit{Gloss-encoder} denoted as \(B_G\) embeds the gloss \(g\) of sense \(s\). The token [CLS] has been proven effective to represent global semantic information. We extract the token [CLS] as representation \(r_g\) from embedding \(E_G\) generated by BERT model

\begin{equation}
    r_g = E_G[0]=B_G(g)[0].
    \label{eq:gloss}
\end{equation}

To match the dimension of \(r_{w^t}^F\), we replicate the \(r_g\) \(poly_m\) times to generate the semantic representation of gloss

\begin{equation}
    r_{g}^F = \mathbf{1}_{poly_m} \otimes r_{g}.
\end{equation}

\subsection{Batch contrastive pre-learning}\label{sec:3.3}
\begin{figure*}[!t]
\centering
\includegraphics[width=\textwidth]{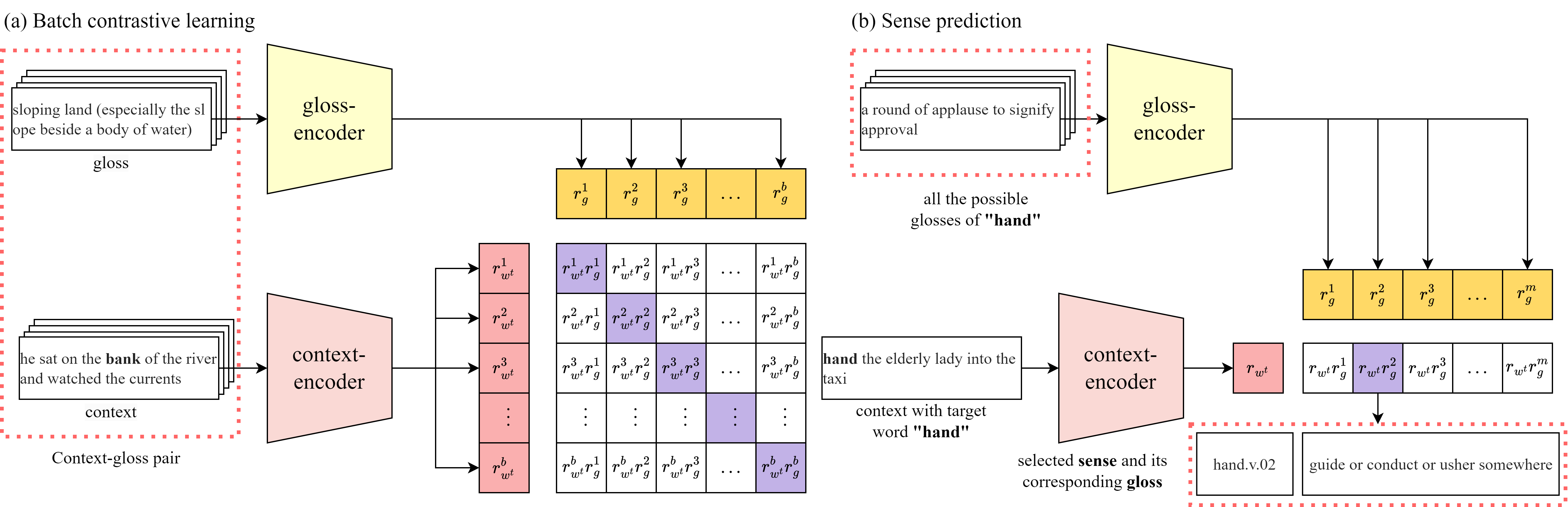}  
\caption{Fusion process is various in two different phase: (a) batch contrastive pre-learning and (b) sense prediction.  }
\label{fig:fusion}
\end{figure*}

PolyBERT fuses representations of target word and gloss after they are generated, the fusion process is shown in Fig. \ref{fig:fusion}.

In the pre-training phase, representations for the target word, \(R_{w^t}=r_{w^t}^1,r_{w^t}^2,\dots ,r_{w^t}^b\) and for glosses, \(R_g=r_g^1,r_g^2,\dots ,r_g^b\) are generated in a batch, where \(b\) denotes the batch size. For each \(r_{w^t}^i\), \(r_g^i\) represents its corresponding gloss's representation. PolyBERT take dot product to fuse two representations in a batch to generate fusion matrix \(M_F\)
\begin{equation}
    M_F=R_{w^t}\cdot R_g,
    \label{eq:repre}
\end{equation}in this matrix, values located on the diagonal line represent the similarities (scores) between target words and the glosses of their correct senses. The other values represent scores between target words and the glosses of incorrect senses. This approach enables PolyBERT take correct senses of other target words in a same batch as incorrect senses (negative samples), which avoids redundant data inputs.

To train the PolyBERT, a special-designed loss function is constructed. Initially, softmax function is applied to normalize the scores of each row of \(M_F\), the process read as
\begin{equation}
    P = \text{softmax}\left(M_{F}\right),
    \label{eq:softmax}
\end{equation}where \(i\) is index of row. Subsequently, normalized scores between \(r_{w^t}^i\) and \(r_g^i\) from \(P\) are extracted:
\begin{equation}
    P^d = \text{diagonal}(P)=\{P_{i,i}|i=1,2,\dots ,b\}.
    \label{eq:extraction}
\end{equation}Based on \(P^d\), the final loss \(\mathbb{L}\) is evaluated
\begin{equation}
    \mathbb{L}= \frac{1}{b}\times \sum_{i=1}^b \left(-\log (P_i^d)\right),
    \label{eq:loss}
\end{equation}where \(i\) is the index of each row. During the pre-training phase, PolyBERT is able to maximize the scores between target words and their corresponding glosses and minimize the scores between target words and their incorrect glosses.

\subsection{Prediction phase}\label{sec:3.4}
In prediction phase, trained PolyBERT is utilized to select the correct sense \(s^c\) and associated gloss \(g^c\) of target word \(w^T\) from candidate senses set \(S_{w^t}\). Initially, representations of target word and candidate senses are generated, the process read as
\begin{equation}
    r_{w^t} = B_C(w^t),
\end{equation}
\begin{equation}
    R_g = B_G(G_{w^t})=r_g^1,r_g^2,\dots ,r_g^m,
\end{equation}where \(G_{w^t}\) is gloss set associated with \(S_{w^t}\) and \(m\) is number of possible senses. PolyBERT employ dot product to calculate scores of each sense:
\begin{equation}
    \mathbb{S}\left(w^t,s^j\right)= (r_{w^t}\cdot R_g)[j],
\end{equation}where \(j\) is index of sense. According to scores of \(s^j \in S_{w^t}\) we can select the most suitable sense \(s^{w^t}\) and its corresponding definition \(g^{w^t}\).

\section{Experiments and Performance evaluation}\label{section:Experiment}
This paper conducts two experiments to evaluate performance and computational costs of PolyBERT. \textbf{Experiment-A} aims to compare performance of PolyBERT against prior works. \textbf{Experiment-B} aims to evaluate whether batch contrastive learning reduce computational cost during training phase.

Experiments are conducted in Google cloud server, which integrated with NVIDIA A100 GPU (RAM is 40GB). The operating system is Ubuntu, version of Python is 3.10.0. Deep learning framework is PyTorch 2,3,1+cu121, with CUDA 12.2. In addition,  the implementation of the model relies on the Transformers library, version 3.14.0.

This section illustrates setups and results of Experiment-A and Experiment-B in section \ref{sec:ex1}, \ref{sec:ex2} respectively. 

\subsection{Experiment-A: performance evaluation}\label{sec:ex1}
Encoders of PolyBERT are initialled with BERT-Large model in Experiment-A. Therefore, dimension of hidden layer is 1024. For \textit{context-encoder} the heads number of multi-heads attention is 8. PolyBERT is trained on SemCor 3.0, which is a large corpus manually annotated with senses from WordNet. Similar to prior works, PolyBERT takes SemEval-2007 (\textbf{SE7}) as develop set, performance of PolyBERT is evaluated in Senseval-2 (\textbf{SE2}), Senseval-3 (\textbf{SE3}), SemEval-2013 (\textbf{SE13}), and SemEval (\textbf{SE15}). Each definition (gloss) of sense is retrieved from WordNet.

\begin{table*}[t]
\centering
\caption{Performances of PolyBERT and carious categories of prior works. In this evaluation, F1-score is taken as indicator of performance. }
\scalebox{1.3}{
\begin{tabular}{|l|c|c|c|c|c|c|c|c|c|c|}
\hline
\textbf{Works}&\multicolumn{1}{c|}{\textbf{Dev}}  & \multicolumn{4}{c|}{\textbf{Test Datasets}} & \multicolumn{5}{c|}{\textbf{Different POS of Test Datasets}} \\ \cline{2-11} 
  &SE7 & SE2 & SE3 & SE13 & SE15  & Nouns & Verbs & Adj. & Adv. & All \\ \hline
MFS &54.5 & 65.6 & 66.0 & 63.8 & 67.1 & 67.7 & 49.8 & 73.1 & 80.5 & 65.5 \\
WordNet S1 & 55.2 & 66.8 & 66.2 & 63.0 & 67.8 & 67.6 & 50.3 & 74.3 & 80.9 & 65.2 \\ 
Basile-LESK\cite{basile2014enhanced} & 56.7 & 63.0 & 63.7 & 66.2 & 64.6  & 70.0 & 51.1 & 51.7 & 80.6 & 64.2 \\ \hline
BiLSTM-K \cite{kaageback2016word} & - & 71.1 &  68.4 & 64.8 & 68.3 & 69.5 & 55.9 & 76.2 &  82.4 & 68.4 \\ 
BiLSTM-R \cite{raganato2017neural} & 64.8 & 72.0 &  69.1 &  66.9 & 71.5 & 71.5 & 57.5 & 75.0 & 83.8 & 69.9 \\ 
HCAN \cite{luo2018leveraging} & - & 72.8 & 70.3 & 68.5 & 72.8  & 72.7 & 58.2 &  77.4 & 84.1 & 71.1 \\ 
EWISE \cite{kumar2019zero} & 67.3 & 73.8 & 71.1 & 69.4 & 74.5 & 74.0 & 60.2 & 78.0 & 82.1 & 71.8 \\ \hline
GLU \cite{hadiwinoto2019improved} &68.1 &75.5 &73.6 &71.1 &76.2 &- &- &- &- &74.1 \\
SVC \cite{vial2019sense} &- &- &- &- &- &- &- &- &- &75.6 \\
ARES \cite{scarlini2020more} &71.0 &78.0 &77.1  &77.3 &83.2 &80.6 &68.3 &80.5 & 83.5 &77.9 \\
GlossBERT \cite{huang2019glossbert} & 72.5 & 77.7 & 75.2 & 76.1 & 80.4 & 79.8 & 67.1 & 79.6 & 87.4 & 77.0 \\
BEM \cite{blevins2020moving} & 74.5 & 79.4 & 77.4 & 79.7 & 81.7 & 81.4 & 68.5 & 83.0 & 87.9 & 79.0 \\ \hline
PolyBERT &\textbf{76.8} &\textbf{81.7} &\textbf{79.6} &\textbf{81.3} &\textbf{83.9} &\textbf{84.6} &68.5 &\textbf{86.7} &\textbf{88.3} &\textbf{81.0} \\ \hline

\end{tabular}
}

\label{table:result-A}
\end{table*}

Table. \ref{table:result-A} (denoted as table in Section \ref{sec:ex1}) shows the results of Experiment A. Experiment A compares the performance of PolyBERT against prior works, using the F1-score to evaluate performance. Prior baseline works are divided into three categories: (1) \textbf{Knowledge-based works}, (2) \textbf{Neural networks-based works}, and (3) \textbf{PLMs-based works}.

The first block of table displays baselines of Knowledge-based works. MFS refers to selecting the most frequent sense for each target word according to the training corpus. WordNet S1 involves selecting the first sense in WordNet, which is the most common sense. Basile-LESK \cite{basile2014enhanced} integrates word embeddings to calculate the degree of overlap between the target word and gloss, representing a variant of the LESK algorithm.

The second block of table outlines baselines of Neural networks-based works. BiLSTM-K \cite{kaageback2016word} uses independent classifiers to disambiguate word senses based on semantics extracted by BiLSTM. BiLSTM-R \cite{raganato2017neural} employs self-attention to enhance the representation of semantics. HCAN \cite{luo2018leveraging} utilizes sense-gloss pairs as external inputs. EWISE \cite{kumar2019zero} pretrains an LSTM-based encoder to generate embeddings from the graph structure of WordNet.

The third block of table presents baselines of PLMs-based works. GLU \cite{hadiwinoto2019improved} uses a gated linear unit to project context and senses into a shared embedding space. SVC \cite{vial2019sense} compresses senses into one candidate to enhance performance. ARES \cite{scarlini2020more} selects the nearest sense based on embeddings of context and glosses. GlossBERT and BEM, as two mainstream approaches, are introduced in Section \ref{section:Related Works}.

The last block of table reports the performance of PolyBERT. The results show that PolyBERT, with an F1-score of 81.0 in all-words WSD, outperforms other works across all evaluation datasets. Moreover, outperforming BEM and GlossBERT highlights the benefits of balanced local and global semantic representation in WSD.

\subsection{Experiment-B: ablation study of batch contrastive learning}\label{sec:ex2}
Experiment-B ablates batch contrastive learning (denoted as BCPL) of PolyBERT in order to verify whether BCPL reduces computational cost. Because poly-encoder will impact processing efficiency, Experiment-B includes bi-encoder based model BEM as control group to isolate impact of poly-encoder. Therefore, four models are trained to be compared: PolyBERT, PolyBERT-A, BEM-C and BEM, where PolyBERT-A refers PolyBERT without BCPL and BEM-C refers to BEM with BCPL. Four models are trained on SemCor corpus and epochs number are 5. The metric used to measure computational cost is GPU hours, calculated using the following formula:
\begin{equation}
    \text{GPU hours}=N \times duration,
    \label{eq:GPU}
\end{equation}where \(N\) is number of GPUs and \(duration\) is time consumption of model training.
\begin{figure}
    \centering
    \includegraphics[width=\linewidth]{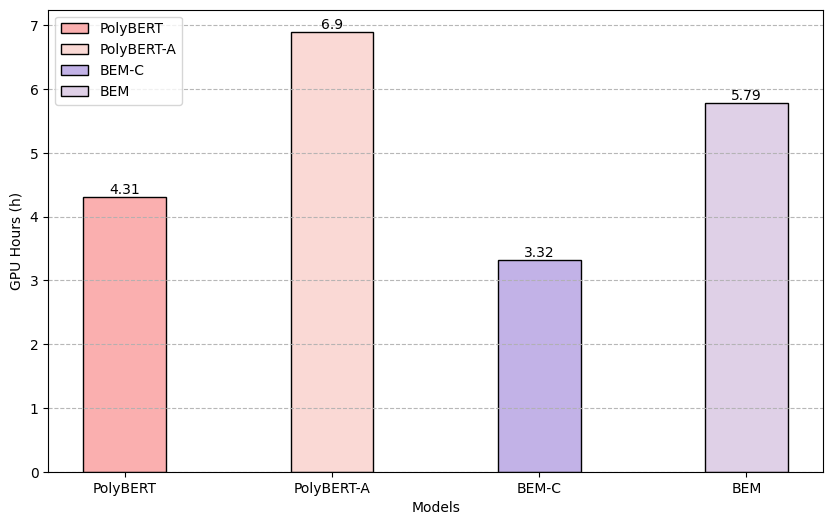}
    \caption{Result of Experiment-B. Four models are trained on SemCor corpus.}
    \label{fig:experiment-B}
\end{figure}

The results, as shown in Figure. \ref{fig:experiment-B}, indicate that \textbf{PolyBERT-A} utilized the most GPU hours, requiring 6.9 GPU hours to complete training. In contrast, PolyBERT consumed 37.6\% fewer GPU hours than PolyBERT-A, demonstrating that BCPL contributes to reducing computational costs. For the control groups, BEM and BEM-C recorded GPU hours of 5.79 and 3.32, respectively, revealing a significant reduction in computational costs for the BEM model when enhanced with BCPL. Overall, these results highlight the effectiveness of BCPL in lowering computational expenses. Additionally, PolyBERT achieves a higher F1-score compared to the other models, further underscoring its superior performance.

\section{Conclusion}\label{sec:conclusion}
This paper introduces a poly-encoder BERT-based WSD model (PolyBERT). PolyBERT addresses two key challenges of mainstream WSD works: (1) imbalanced representation of local and global semantics and (2) unnecessary computational costs associated with redundant training data inputs. To address challenge (1), PolyBERT employs a poly-encoder with a multi-head attention mechanism to effectively fuse token-level (local) and sequence-level (global) semantics. To address challenge (2), PolyBERT incorporates batch contrastive learning (BCL) during the training phase, which enables the model to be trained using only positive samples, thereby reducing computational overhead.

The primary innovations of PolyBERT lie in two aspects: first, the use of a poly-encoder to integrate token-level and sequence-level semantics, which enriches the model's semantic representation by balancing local and global information; and second, the introduction of batch contrastive learning, which significantly reduces training costs by eliminating the need for negative samples. Based on these contributions, PolyBERT not only improves performance but also reduces computational costs. Experimental results demonstrate that PolyBERT outperforms mainstream models, achieving a 2\% higher F1-score. In addition, by employing batch contrastive learning, PolyBERT reduces GPU hours by 37.6\% during the training phase.

However, like mainstream works, PolyBERT still requires a large amount of manually labeled data as a training set to achieve optimal performance. This underscores a primary direction for future WSD research: enhancing few-shot approaches. Few-shot WSD approaches will enable the adaptation of WSD systems across various domains, which is crucial in certain fields of expertise.

\bibliographystyle{IEEEtran}
\bibliography{reference}

\end{document}